\documentclass[letterpaper, conference]{ieeeconf}
\IEEEoverridecommandlockouts
\usepackage{cite}

\def\BibTeX{{\rm B\kern-.05em{\sc i\kern-.025em b}\kern-.08em
    T\kern-.1667em\lower.7ex\hbox{E}\kern-.125emX}}
    
  \usepackage{amsmath,amssymb,amsfonts,amsthm} 
\usepackage{mathtools}
\usepackage{commath}
\usepackage{algorithmic}
\usepackage[ruled,vlined]{algorithm2e}
\usepackage{graphicx}
\usepackage{textcomp}
\usepackage[table]{xcolor}

\usepackage{framed}
\definecolor{shadecolor}{rgb}{.9,.9,.9}
\usepackage{tcolorbox}
\usepackage{siunitx}
\usepackage{textgreek}
\usepackage{url}
\usepackage{booktabs, tabularx}
\usepackage{multirow}
\usepackage{stackengine}
\usepackage[caption=false, font=footnotesize]{subfig}

\usepackage{tikz}
\usetikzlibrary{quotes,angles}
\usetikzlibrary{calc}
\usetikzlibrary{decorations.pathmorphing}

\usepackage{soul}
\usepackage{todonotes}

\usepackage[absolute]{textpos}     
\usepackage[hidelinks]{hyperref}   

\title{\LARGE\bf Thrust Direction Control of an Underactuated Oscillating Swimming Robot
}

\author{Gedaliah Knizhnik and Mark Yim
\thanks{The authors are with the GRASP Laboratory, University of Pensylvannia, Philadelphia, PA 19104. 
        {\tt\footnotesize knizhnik@seas.upenn.edu}}%
}

\newcommand{\copyrightstatement}{
    \begin{textblock*}{5.7in}(0.25in,0.25in) 

        \noindent
        \footnotesize
        This accepted article to IROS is made available by the authors in compliance with IEEE policy.

        \noindent
        Please find the final, published version in IEEE Xplore, DOI: \href{https://doi.org/10.1109/IROS51168.2021.9636778}{\textcolor{blue}{https://doi.org/10.1109/IROS51168.2021.9636778}}.
        
    \end{textblock*}

    \begin{textblock*}{5.7in}[0,1](0.25in,10.85in) 

        \noindent
        \footnotesize \scriptsize
        \copyright 2021 IEEE. Personal use of this material is permitted.
        Permission from IEEE must be obtained for all other uses, in any current or future media, including reprinting/republishing this material for advertising or promotional purposes, creating new collective works, for resale or redistribution to servers or lists, or reuse of any copyrighted component of this work in other works.
    \end{textblock*}
}

\DeclareMathOperator{\sign}{sgn}

\begin{document}
\bstctlcite{MyBSTcontrol} 
\copyrightstatement                    

\maketitle

\begin{abstract}
The Modboat is an autonomous surface robot that turns the oscillation of a single motor into a controlled paddling motion through passive flippers. Inertial control methods developed in prior work can successfully drive the Modboat along trajectories and enable docking to neighboring modules, but have a non-constant cycle time and cannot react to dynamic environments. In this work we present a thrust direction control method for the Modboat that significantly improves the time-response of the system and increases the accuracy with which it can be controlled. We experimentally demonstrate that this method can be used to perform more compact maneuvers than prior methods or comparable robots can. We also present an extension to the controller that solves the reaction wheel problem of unbounded actuator velocity, and show that it further improves performance.
\end{abstract}


\section{Introduction} \label{sec:intro}

The Modboat is a single-motor surface swimmer composed of two roughly cylindrical bodies linked by a motor, with passive flippers mounted on the bottom body. Oscillating the motor causes the flippers to open sequentially, resulting in a steerable paddling motion. This creates a simple system that can --- if properly controlled --- serve as an oceanographic sensing platform for both coarse and fine-resolution data collection. Actuators tend to dominate cost, so the Modboat's single-motor design enables building large groups for extensive data gathering at a much lower cost than traditional approaches.

In prior work we presented two discrete controllers that could steer the Modboat using either differential thrust (by modifying the frequency of oscillation) or inertial rotation (by inserting pauses into the oscillation)~\cite{Knizhnik2020PausesRobot}. Differential thrust has a constant cycle time, which is good for planning, but low control authority, resulting in slow response. Inertial control has more authority but a non-constant cycle time; this makes coordinating tasks or responding to dynamic environments challenging. Precision oceanography, such as monitoring specific animals or inspecting infrastructure, however, requires both high control authority for maneuvering and a constant cycle time for planning and coordinating.

Moreover, individual Modboats can dock together into a square lattice to enable fine-resolution data collection, which conventional boats cannot do. In prior work we showed that docking is achievable using discrete inertial control despite its challenges~\cite{Knizhnik2021DockingRobot}; but we were successful only in still water, whereas disturbances are frequent under real conditions. Inertial control cannot react sufficiently quickly to such disturbances, and the non-constant cycle time complicates planning for them. We need a control method that has both high authority and a constant cycle time to enable more robust docking, but designing such control for a non-standard design such as the Modboat is difficult.


Control laws for such non-standard designs have been explored for foil-shaped swimming robots by Pollard, Fedonyuk, and Tallapragada \cite{Pollard2017AnRotor,Pollard2019SwimmingConstraints,Fedonyuk2020DynamicsFreedom} and Lee, Ghanem, and Paley \cite{Lee2019State-feedbackVehicle}\cite{Ghanem2020PlanarFish}. These systems leverage a no-slip condition, oscillation of a reaction mass, and vortex shedding to generate propulsion~\cite{Tallapragada2015} and can demonstrate complex and coordinated behaviors. But the same no-slip conditions that define them also significantly reduce their maneuverability. 

Relaxing the no-slip constraint, Pollard used a small-slip condition at the foil tip instead, but this did not change the control methods or increase maneuverability~\cite{Pollard2019SwimmingConstraints}. Dear et al. considered lateral skidding as a low-magnitude constraint violation for a (terrestrial) snakeboard system~\cite{Dear2015SnakeboardSkidding}, whose dynamics are defined by two no-slip conditions, and saw some success in generating sharper turns. Significant increases in maneuverability, however, can result from constructing systems without these constraints, rather than violating them.

In this work we show that similar oscillatory control laws can propel the Modboat through existing paddling dynamics without a no-slip condition. We also demonstrate that --- once we compensate for the lack of no-slip condition --- the resulting system is more maneuverable and significantly outperforms both foil-shaped robots and previous Modboat controllers.

\begin{figure}[t]
    \centering
    \includegraphics[trim = 0cm 0.1cm 0cm 0.02cm, clip=true, width=0.95\linewidth]{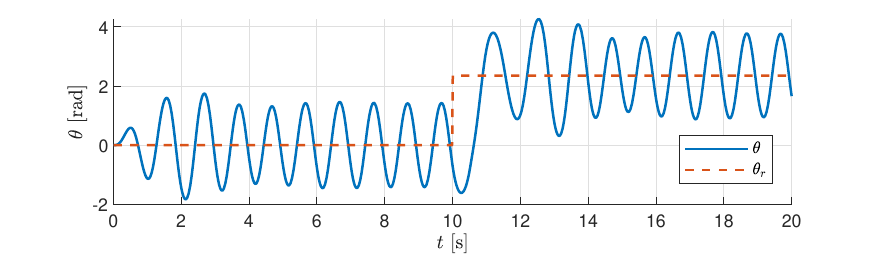}
    \caption{Simulated convergence of~\eqref{eq:eomMain} to oscillation about a reference heading under control law~\eqref{eq:contLimCyc}. In this simulation $C_f=\num{1.0e-4}$, $C_r = 0$, $I_b = \num{5.2e-6}$, and $I_t = \num{1.0e-3}$. For the controller, $\omega = 2\pi$, $K = 15$, and $\beta = 40$.}
    \label{fig:simConv}
\end{figure}

This paper is organized as follows. Sec.~\ref{sec:dynamics} describes Modboat rotational dynamics and presents the control law. Sec.~\ref{sec:travelDir} presents the adjustment to the control law made to control the direction of travel without a no-slip condition, which forms our novel \textbf{thrust direction} controller. Sec.~\ref{sec:desaturating} introduces a method for reducing the angular velocity of the reaction mass, which forms the novel \textbf{desaturated thrust direction} controller. Finally, Secs.~\ref{sec:experiments} and~\ref{sec:discussion} present experimental verification and discussion, respectively.


\section{Dynamics} \label{sec:dynamics}

The planar equations of motion for the Modboat have been presented in detail in prior work~\cite{Knizhnik2020}. The translational dynamics of the Modboat are complex, so we omit them here. We can analyze its movement approximately, however, by considering only the orientation. The translational dynamics of the Modboat under a symmetric control law can then be approximated as a thrust along the average orientation.

The orientation equation is reproduced in~\eqref{eq:eomMain}, where $\theta$ is the orientaton of the Modboat's bottom body (which contains the flippers), $I_b$ and $I_t$ are the moments of inertia for the bottom and top body, respectively, and $\phi$ is the motor angle. $C_f$ and $C_r$ are drag constants related to the flippers and the cylindrical shape respectively. The top body acts as a reaction wheel to transfer angular momentum to the bottom body.
\begin{equation} \label{eq:eomMain}
(I_b + I_t) \ddot{\theta} = -C_f \sign{(\dot{\theta})} \dot{\theta}^2 - C_r \dot{\theta} - I_t \ddot{\phi}
\end{equation}

\subsection{Control to a Limit Cycle}

The propulsive mechanism for foil-shaped robots is vortex shedding off the foil tip~\cite{Tallapragada2015}. The Modboat is propelled by flippers rather than a foil, but it is reasonable to think a similar control law may be effective. Oscillations around a reference heading should cause paddling in that direction, just as a foil would create thrust. 

A number of oscillating control laws have been considered for foil-shaped robots. Pollard et al. used a sinusoidal forcing term $\sin{(\omega t)}$ and an integral term $\frac{1}{T} \int_{t-T}^t (\theta_r - \theta)dt$ for convergence~\cite{Pollard2019SwimmingConstraints} and showed that it could drive a robot to a desired velocity and reference heading on average. Lee et al. used an angular velocity term and a sinusoidal convergence term $\sin{(\theta_r - \theta)}$~\cite{Lee2019State-feedbackVehicle}, but did not significantly characterize its performance. Fedonyuk used a control law of the form in~\eqref{eq:contLimCyc} for a terrestrial analogue, and this is the form we will consider\footnote{The full form of~\eqref{eq:contLimCyc} used in~\cite{Fedonyuk2020DynamicsFreedom} includes a constant $\tau_0$ that is used for curved trajectories. We do not consider curving trajectories in this work.}~\cite{Fedonyuk2020DynamicsFreedom}.
\begin{equation}\label{eq:contLimCyc}
    \tau = -K \sin{(\omega t)} - \beta \sin{\left (\theta_r(t) - \theta(t) \right )}
\end{equation}

In~\eqref{eq:contLimCyc}, $\sin{(\omega t)}$ drives the orientation at the angular frequency $\omega$, while $\sin{(\theta_r - \theta)}$ serves to drive the orientation $\theta$ to the reference heading $\theta_r$ on average. Let $\ddot{\phi}$ in~\eqref{eq:eomMain} by given by $\tau$ in~\eqref{eq:contLimCyc} and substitute $\psi = \theta - \theta_r$. Then the resulting equation of motion~\eqref{eq:pendulum} is a \textit{nonlinear damped pendulum} oscillating around $\psi = 0$, with $D(\dot{\psi})$ collecting the linear and nonlinear damping terms.
\begin{equation} \label{eq:pendulum}
    \ddot{\psi} + D(\dot{\psi}) + \frac{\beta I_t}{I_t + I_b} \sin\left ( \psi \right ) = \frac{K I_t}{I_t + I_b} \sin(\omega t)
\end{equation}

In~\eqref{eq:pendulum}, we see that while $K\sin{(\omega t)}$ acts as a forcing function at angular frequency $\omega$, the second control term $\beta\sin{(\psi)}$ acts as part of the dynamics. In particular, its coefficient is related to the natural frequency of the pendulum. To ensure a stable oscillation the forcing frequency and the natural frequency should be aligned as in~\eqref{eq:omega}.

\begin{equation} \label{eq:omega}
    \omega^2 \approx \frac{\beta I_t}{I_t + I_b} 
\end{equation}

In practice $I_t \gg I_b$ to allow most of the rotation to transfer to the bottom body. Moreover we do not need perfect resonance --- just enough alignment to avoid large oscillations in amplitude --- so it is sufficient to say $\beta \approx \omega^2$.

The controller in~\eqref{eq:contLimCyc} therefore has two design variables. The angular frequency $\omega$ controls the paddling frequency and therefore the controller responsiveness. $K$, meanwhile, loosely controls the oscillation amplitude, and functions as a design variable to set forward velocity relative to a given angular frequency $\omega$. We have a constant cycle time because $\omega$ is generally constant, and we will show in Sec.~\ref{sec:experiments} that this controller has significant control authority.

\subsection{Simulation}

We can verify in simulation that this control law produces convergence to a desired heading. Numerically solving~\eqref{eq:eomMain} and~\eqref{eq:contLimCyc} shows that orientation converges to a limit cycle around the desired reference heading as long as there is sufficient drag\footnote{We leave the analysis of how significant the drag terms must be to future work. For the current analysis, it is sufficient to note that there is enough drag in water, but not in air.} as shown in Fig.~\ref{fig:simConv}. Note that there is a transient region at the discontinuity in reference heading during which the system experiences net torques. The extent of this region depends on the phase of the transition time, but its effect on the overall trajectory is negligible. 

\begin{figure}[t]
    \centering
    \includegraphics[width=\linewidth]{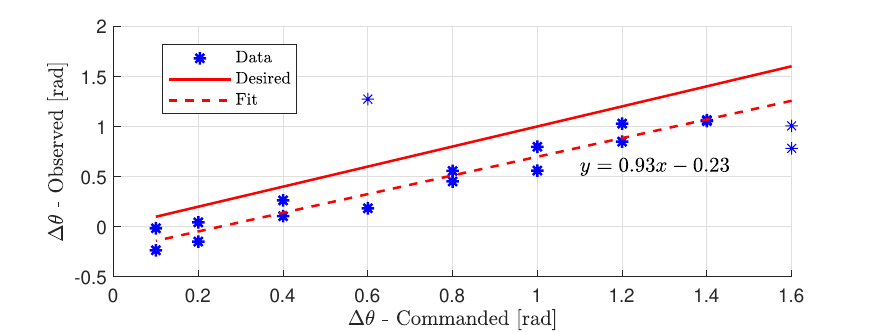}
    \caption{The observed direction of travel is compared to the commanded change in reference heading under the control law~\eqref{eq:contLimCyc}. The observed change consistently lags the desired change due to drift.}
    \label{fig:stepResponses}
\end{figure}


\section{Travel Direction} \label{sec:travelDir}

The foil-shaped robots developed by Pollard et al.~\cite{Pollard2019SwimmingConstraints}, Fedonyuk~\cite{Fedonyuk2020DynamicsFreedom}, and Lee et al.\cite{Lee2019State-feedbackVehicle} use either a knife-edge constraint (Kutta condition) or a small periodic slip velocity constraint on the foil tip. This equates the average orientation and the direction of travel, so~\eqref{eq:contLimCyc} is sufficient to effectively drive the robot in the plane. 

The Modboat does have a foil-shaped tail~\cite{Knizhnik2021DockingRobot}\cite{Knizhnik2020}, but its negligible thickness means that neither a knife-edge constraint nor a small slip velocity constraint can be applied to it. This means drift is a significant component of motion, and~\eqref{eq:contLimCyc} is not sufficient to drive the Modboat in the plane. This is shown in Fig.~\ref{fig:stepResponses}, where the observed change in the trajectory direction under a step input to~\eqref{eq:contLimCyc} is consistently about $0.2\si{rad}$ lower than the commanded change in reference heading under~\eqref{eq:contLimCyc}. This is not a large error, but it is compounded every time the reference heading changes.

This behavior is consistent with a simple thrust model, in which thrust is along the direction of the reference heading $\theta_r$ (balanced by drag). Changing the direction of thrust does not immediately cancel the previous velocity (like friction would given a knife-edge constraint), and the net direction of travel is a vector sum of the old and new thrusts. The trajectory will converge to the new thrust direction eventually, but this is not a feasible approach for precise maneuvering.

\begin{figure}[t]
    \centering
    \includegraphics[trim = 0cm 0.1cm 0cm 0.2cm, clip=true, width=\linewidth]{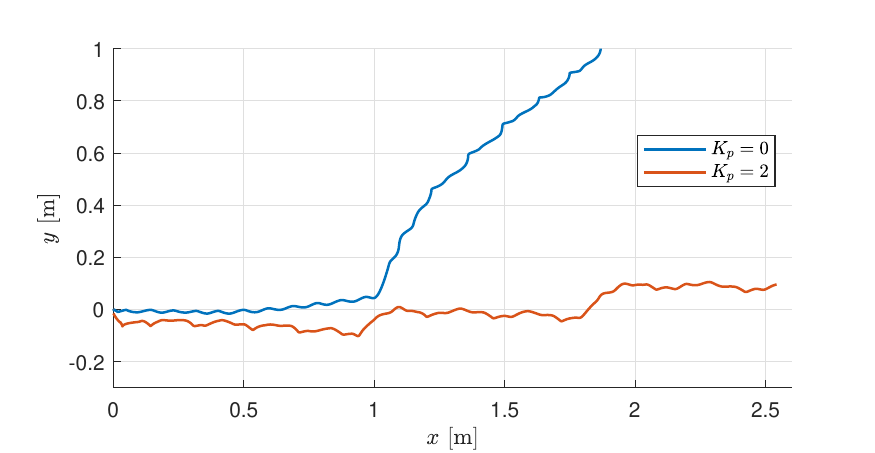}
    \caption{Trajectories under a constant reference heading of $0\si{rad}$. When using the control law~\eqref{eq:contLimCyc} alone (blue), even a single disturbance (applied at $\approx (1,0)$) is enough to decouple the reference heading and the direction of travel. Adding the control law~\eqref{eq:contDirTrav} (orange line) allows the direction of travel to remain in the face of multiple disturbances. Note that both trajectories have been rotated to facilitate comparison.}
    \label{fig:contComp}
\end{figure}

Note, the control laws presented in this section use vectors. For clarity, we do not include normalization steps. Instead we use $\hat{v}$ to denote the unit vector of $\vec{v}$, and where relevant equalities should be understood with an implicit normalization.


\subsection{Controlling the Direction of Travel}

To control the direction of travel we use an additional control layer~\eqref{eq:contDirTrav}, where $\hat{\theta}$ indicates a unit vector oriented at an angle $\theta$ relative to the $x$-axis. $\hat{\theta}_{des}$ is then the feed-forward desired direction of travel, $\hat{\psi}$ is the observed direction of travel, and $K_p$ is a proportionality constant.

\begin{equation} \label{eq:contDirTrav}
\hat{\theta}_{r}(t) = \hat{\theta}_{des}(t) + K_p \left (\hat{\theta}_{des}(t) - \hat{\psi}(t) \right )
\end{equation}

Eq.~\eqref{eq:contDirTrav} acts as a proportional controller on the thrust direction to generate the desired direction of travel; it cancels disturbances or existing velocity when changing direction. External flows can also be included in $\hat{\psi}$ if they are known. The vector form of~\eqref{eq:contDirTrav} ensures reasonable outputs regardless of error magnitude or proportionality constant. Together,~\eqref{eq:contLimCyc} and~\eqref{eq:contDirTrav} constitute our novel \textbf{thrust direction} controller.

Fig.~\ref{fig:contComp} shows the improvement that results from using~\eqref{eq:contDirTrav} in addition to~\eqref{eq:contLimCyc}. Under~\eqref{eq:contLimCyc} alone, even a single disturbance decouples $\theta_r$ and the direction of travel for significant time. But under~\eqref{eq:contDirTrav} as well multiple disturbances can be quickly rejected due to the observed change in direction of travel. Eq.~\eqref{eq:contDirTrav} causes the paddling direction to cancel the disturbance and return to the desired heading. A similar behavior corrects the step-input performance in Fig.~\ref{fig:stepResponses}.

All the terms in~\eqref{eq:contLimCyc} and~\eqref{eq:contDirTrav} are defined as continuous functions of time. Although thrust is still produced over complete cycles, this allows this controller to respond faster than the  discrete controllers in our prior work~\cite{Knizhnik2020PausesRobot}.


\subsection{Measuring the Direction of Travel}

Because the Modboat oscillates and rocks as it moves, it is difficult to quantify the direction of travel $\hat{\psi}(t)$. Direct differentiation of the trajectory would yield significant oscillation, which is undesirable when used in~\eqref{eq:contDirTrav}. Let $\vec{x}(t) = [\begin{matrix} x(t) & y(t) \end{matrix}]'$ be the trajectory of the Modboat, and let $T$ be the period of oscillation associated with the angular frequency $\omega$. Then~\eqref{eq:periodVel} represents the \textit{period-wise velocity}, which provides a smoother measure of motion than the instantaneous velocity. We obtain $\hat{\psi}(t)$ by taking the orientation of $\vec{v}_T(t)$ and then averaging over the prior period $T$ as in~\eqref{eq:travelDir}. This averaging has been heuristically determined to provide sufficient smoothness for good performance of~\eqref{eq:contDirTrav}.

\begin{equation} \label{eq:periodVel}
    \vec{v}_T(t) = \begin{bmatrix} \dot{x}_T(t) \\ \dot{y}_T(t) \end{bmatrix} = \frac{\vec{x}(t) - \vec{x}(t-T)}{T} 
\end{equation}




\begin{equation} \label{eq:travelDir}
    \hat{\psi}(t) = \frac{1}{T} \int_{t-T}^{t} \arctan{\left ( \frac{\dot{y}_T (\tau)}{\dot{x}_T (\tau)} \right )} d\tau
\end{equation}

Use of this methodology requires at least $2T$ of initialization data before $\hat{\psi}(t)$ is well defined. This can be shortened to $T$ by letting the orientation of $\vec{v}_T$ be $\hat{\theta}_{des}(t)$ $\forall t<T$, which is a reasonable assumption in still water. But it is also reasonable to require some startup time before assuming accurate velocity measurements.


\section{Desaturating the Actuator} \label{sec:desaturating}

The controller in~\eqref{eq:contLimCyc} produces no net torque on the system in steady-state when $\theta_{r}(t)$ is held constant. Net torques are generated, however, in the transient regions when $\theta_{r}(t)$ changes. If not properly controlled, these torques lead to increasing angular velocity for the top-body (denoted $\dot{\theta}_t$), which is effectively a reaction wheel in this configuration. This introduces the classic reaction wheel problem in that actuators cannot increase in speed indefinitely, and torques held for long periods (e.g. steady state) are not possible.

\begin{figure}[t]
    \centering
    \includegraphics[trim = 0cm 0.1cm 0cm 0.2cm, clip=true, width=\linewidth]{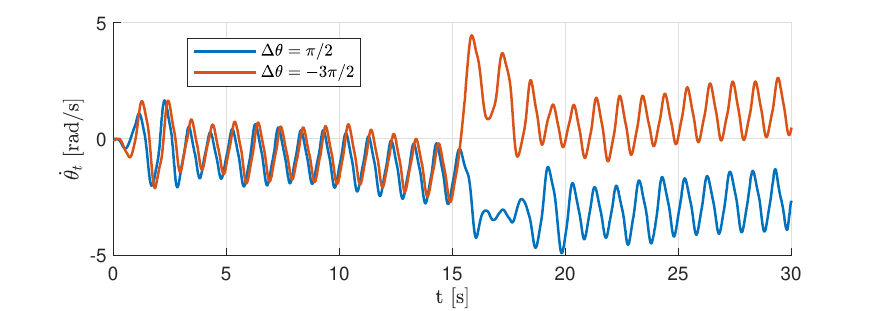}
    \caption{A plot of top-body angular velocity over time for two step input tests. At $t=15\si{s}$ the desired heading is changed by $\pi/2$($-3\pi/2$) for the blue(orange) curve. The final desired heading is the same, but the choice determines the change in angular velocity for the actuator.}
    \label{fig:desatComp}
\end{figure}

In conventional reaction wheel systems a number of methods exist to reduce this effect, but the first step is to choose an actuator with a high saturation ceiling. This works because we are not typically concerned with the state of the reaction body. This is the case for the foil-shaped robots designed by Pollard et al.~\cite{Pollard2019SwimmingConstraints} and Lee et al.~\cite{Lee2019State-feedbackVehicle}, in which the reaction mass is internal. Other solutions can leverage dead-bands in the dynamics; for example, Dear et al. resets the rotor velocity of their terrestrial snakeboard system by bringing it to a full stop, at which point the rotor can be slowed without affecting the system~\cite{Dear2015SnakeboardSkidding}.

The Modboat, however, cannot use these solutions. As discussed~\cite{Knizhnik2020} and demonstrated~\cite{Knizhnik2021DockingRobot} in our prior work, the Modboat top body is used for docking to other modules and houses all sensing structures. To enable docking and effective sensing, the angular velocity of the top body must be kept as low as possible. It is also unreasonable to require a full stop to reduce the velocity if alternatives can be found.

We can, however, desaturate the actuator by leveraging the fact that orientation is cyclic. For any change in heading $\Delta \theta_{r}$, rotating by $2\pi-\Delta \theta_{r}$ produces the same final heading but an opposite transient effect. Some angular velocity must remain for the system to move, but we can maintain an upper bound to the angular velocity at minor cost to cornering.

The controller in~\eqref{eq:contLimCyc} cannot be used this way, as it cannot discriminate between congruent angles $\theta+2\pi n$ $\forall n \in \mathbb{Z}$. We can modify it to achieve this effect by rewriting~\eqref{eq:contLimCyc} into~\eqref{eq:contAdj}, with the additional term $\xi(\theta, \theta_r)$ given by the boolean expression in~\eqref{eq:contBool}. The subscript $(-\pi,\pi]$ indicates an angle wrapping operation, so the boolean expression in~\eqref{eq:contBool} serves to override the convergence term in~\eqref{eq:contAdj} until $\theta$ approaches the correct unwrapped value of $\theta_r$. Then the controller reduces back to~\eqref{eq:contLimCyc}. 
\begin{equation} \label{eq:contAdj}
\tau = - K \sin{(\omega t)} - \beta \left [\sin\left ( \theta_r - \theta \right ) + \xi(\theta, \theta_r) \right ]
\end{equation}
\begin{equation} \label{eq:contBool}
    \xi(\theta, \theta_r) =  \sign(\theta_r - \theta) \left [ \left ( \theta_r - \theta \right )_{(-\pi,\pi]} \neq \left ( \theta_r - \theta \right ) \right ]
\end{equation}

Under~\eqref{eq:contAdj} a reduction in angular velocity is accomplished by adding or subtracting $2\pi$ from the reference heading as per~\eqref{eq:refAdj}. This is shown in Fig.~\ref{fig:desatComp}, where a nearly identical angular velocity is maneuvered to be either more positive or more negative by the choice of the new reference heading. An intelligent choice of reference heading can thus reduce (or maintain) the average magnitude of the angular velocity. 
\begin{equation} \label{eq:refAdj}
\theta_{r} = \theta_{r} \pm 2\pi \sign{(\bar{\dot{\theta}}_t)}
\end{equation}

Note that equation~\eqref{eq:refAdj} must be used intelligently. Positive(negative) changes in $\theta_r$ will cause negative(positive) changes in the angular velocity without additional adjustment. Therefore~\eqref{eq:refAdj} should only be used if $\Delta \theta_r$ does not already contribute in the correct direction, and only at discrete intervals to avoid significant transient events.

Together, the direction of travel controller~\eqref{eq:contDirTrav}, the adjusted limit-cycle controller~\eqref{eq:contAdj}, and the boolean expression~\eqref{eq:contBool} constitute our novel \textbf{desaturated thrust direction} controller. 

\begin{figure}
    \centering
    \includegraphics[trim = 0cm 1.2cm 0cm 1.2cm, clip=true, width=\linewidth]{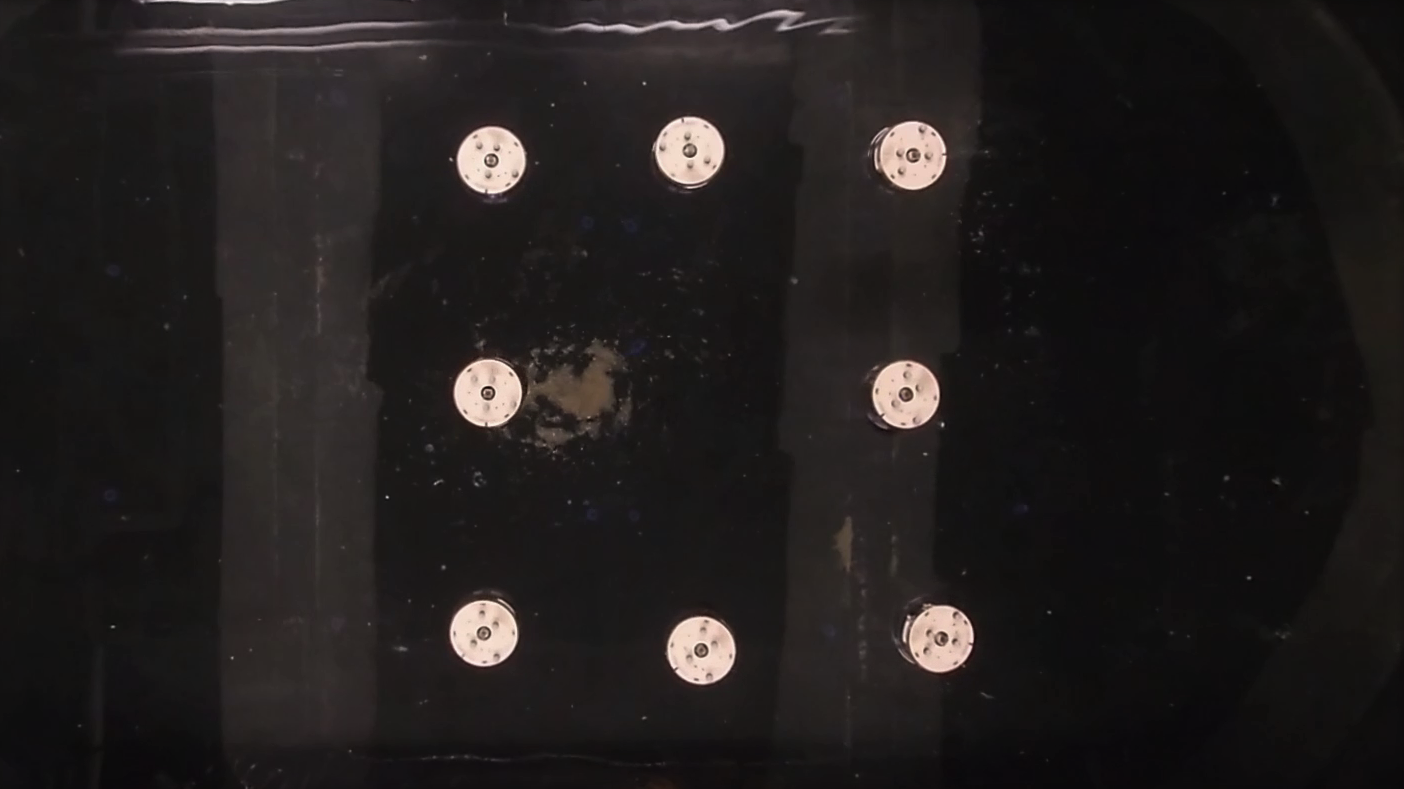}
    \caption{Composite image of the Modboat executing a square trajectory in the testing tank. The associated trajectory plot is shown in Fig.~\ref{fig:trajSquare}.}
    \label{fig:waypointImg}
\end{figure}


\section{Experiments} \label{sec:experiments}

We quantified the Modboat's performance under the thrust direction controller in a $4.5\si{m} \times 3.0\si{m} \times 1.2 \si{m}$ tank of still water (Fig.~\ref{fig:waypointImg}) equipped with an OptiTrack motion capture system that recorded planar position, orientation, and velocities at $120 \si{Hz}$. A MATLAB script recorded the data, calculated the reference heading, and then transmitted the observed and desired orientations to the Modboat over WiFi. Thus the outer control loop~\eqref{eq:contDirTrav} ran off-board at $120 \si{Hz}$, but the inner loop~\eqref{eq:contAdj} ran on-board at $250\si{Hz}$.

\begin{table}[t]
    \centering
    \caption{Control parameter values used in~\eqref{eq:contDirTrav} and~\eqref{eq:contAdj}.}
    \begin{tabularx}{\linewidth}{>{\centering\arraybackslash}X|>{\centering\arraybackslash}X|>{\centering\arraybackslash}X|>{\centering\arraybackslash}X } \toprule 
        $\omega$ $[\si{rad/s}]$ & $K$ $[\si{\newton\meter}]$ & $\beta$ $[\si{\newton\meter}]$ & $K_p$ \\ \midrule
        $2\pi$ & $15$ & $40$ & $1.5$ \\ \bottomrule 
    \end{tabularx}
    \label{tab:controlParams}
\end{table}

The limit cycle control loop~\eqref{eq:contAdj} was evaluated by running step-input tests under the parameters given in Table~\ref{tab:controlParams}.  An initial swim direction for $15\si{s}$ was followed by a change of $\Delta \theta$, and the direction of travel, as measured by~\eqref{eq:travelDir}, was considered for both legs of the trajectory. With proper calibration of the observed orientation and the internal motor zero position the initial travel direction was within a few degrees of the desired value. The results for the second leg are shown in Fig.~\ref{fig:stepResponses}, where the error is $0.2\si{rad} \approx 10^\circ$. $\Delta \theta$ for all tests was positive, so a negative error is consistent with the presented thruster model. 

\begin{figure}[t]
    \centering
    \subfloat[\label{fig:trajSquare}]{ \includegraphics[trim = 0cm 0cm 0cm 0.8cm, clip=true, width=\linewidth]{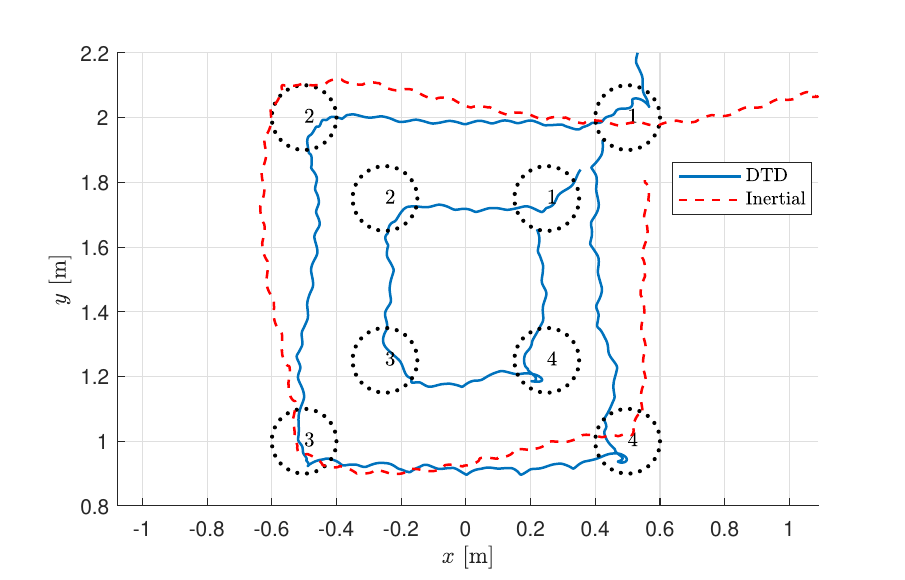}
        }
          \hfill
    \subfloat[\label{fig:trajLine}]{\includegraphics[trim = 0cm 0cm 0cm 0.6cm, clip=true, width=\linewidth]{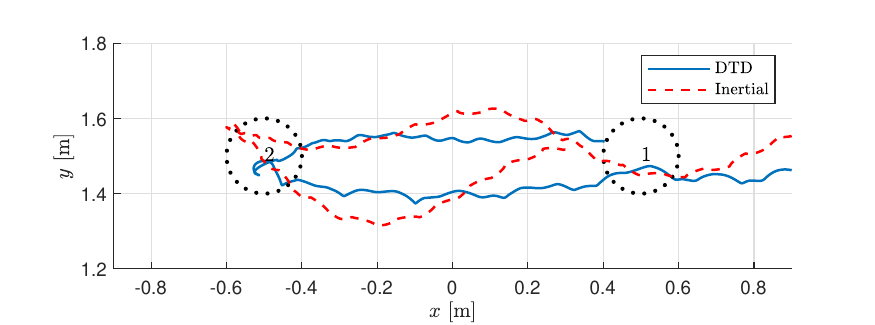}
        }
    \caption{Trajectories of the Modboat following a (a) square waypoint layout (the outer is $1.0\si{m}$ to a side, while the inner is $0.5\si{m}$) and (b) linear waypoint layout (of length $1.0\si{m}$) under desaturated thrust direction (DTD) control (blue solid) and inertial control (red dashed). Displayed waypoints are $0.1\si{m}$ in radius as used by thrust direction control, although the inertial tests used $0.2\si{m}$ radius waypoints. Artifacts of desaturation maneuvers can be observed in (a) both waypoints 4 and (b) waypoint 2. }
    \label{fig:trajs}
\end{figure}

We evaluated the thrust direction controller~\eqref{eq:contDirTrav} and~\eqref{eq:contAdj} by tasking the Modboat to swim to a series of discrete waypoints in sequence under the parameters given in Table~\ref{tab:controlParams}, both with and without the desaturation term~\eqref{eq:contBool}. The waypoints were placed to approximate square, triangular, and linear trajectories, with a tolerance used to determine the transition from one segment to the next. 
Several resulting trajectories are shown in Fig.~\ref{fig:trajs} in comparison with equivalent trajectories from the inertial (pause) controller from our prior work~\cite{Knizhnik2020PausesRobot}. These trajectories demonstrate the overall linearity of the the trajectory segments and the sharpness with which the Modboat can turn under this control law. Maneuvering capabilities extracted from these tests are presented in Table~\ref{tab:comparison}.

\begin{figure}
    \centering
    \includegraphics[trim = 0cm 0.8cm 0cm 0.8cm, clip=true, width=\linewidth]{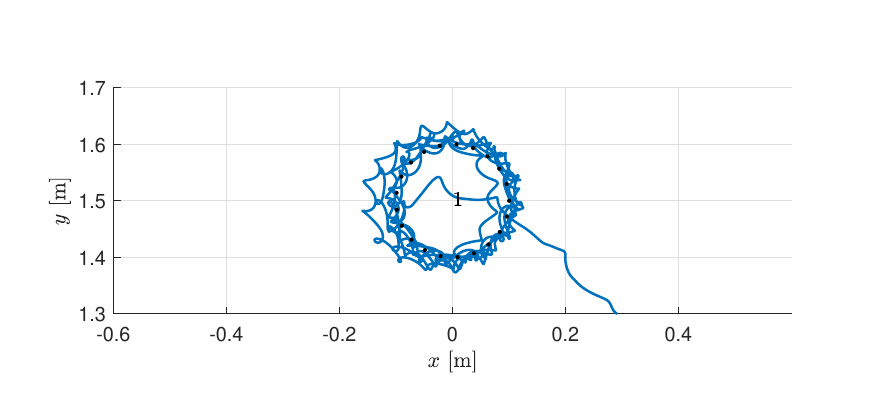}
    \caption{Trajectory of the Modboat commanded to swim towards the waypoint at $(0,1.5)$ under desaturated thrust direction control. The orbit radius is $0.11\si{m}$, which is $1.5$ times the radius of the Modboat.}
    \label{fig:stationKeep}
\end{figure}

Finally, we demonstrated that the existing control laws --- without further modification --- allow the Modboat to execute station-keeping maneuvers. Commanding only one waypoint results in normal swimming behavior that becomes an orbit of radius $0.11\si{m}$ (or $1.5$ times the Modboat radius) around the waypoint, as shown in Fig.~\ref{fig:stationKeep}. 

\section{Discussion} \label{sec:discussion}

Development of thrust direction control in this work was motivated by a need for increased maneuverability and responsiveness in docking applications. We consider two types of maneuvers: $90^\circ$ turns as a proxy for normal maneuvering, and $180^\circ$ turns as a proxy for more extreme maneuvers or abort procedures\footnote{It is --- of course --- possible to do a $180^\circ$ turn by extending the procedure for a $90^\circ$ turn. This is not preferred, however, if there is significant travel during normal turning. Thus Modboat control for $90^\circ$ and $180^\circ$ turning is the same, but for the foil-shaped robots it is different.}. We also consider the minimum size of target that can be achieved by the controller, the minimum distance in which it can navigate to targets, and the RMS perpendicular distance to the ideal straight-line trajectory. 

As summarized on the left side of Table~\ref{tab:comparison}, thrust direction control performs better than our prior techniques on all metrics except RMS error, where it is comparable. It can hit targets of radius $5\si{cm}$ given $1\si{m}$ of distance, and those of radius $10\si{cm}$ at $0.25\si{m}$ (although the trajectories are straighter given $0.5\si{m}$). Since the Modboat radius is $7.5\si{cm}$ with the radius of acceptance extended another $15\si{cm}$ by magnets~\cite{Knizhnik2021DockingRobot}, we can dock successfully using this technique with a margin of error to account for disturbances. 
Higher precision can also be obtained by lowering $K$ to decrease forward velocity and/or raising $K_p$ to increase preference for canceling velocity direction error.

We can increase precision further by using the \textit{desaturated} thrust direction controller to improve trajectory error. This is indicated in Table~\ref{tab:comparison}, as desaturated thrust direction control shows a statistically significant improvement in trajectory error over inertial control, whereas regular thrust direction control does not. The transient effects produced by~\eqref{eq:contBool} are minimal  and do not reduce maneuverability. This is shown in Table~\ref{tab:comparison}, as maneuvering characteristics are not worsened by the addition of desaturation, and are actually improved for abort maneuvers. Minor  artifacts that the desaturation term introduces can be seen in the trajectory corners of Fig.~\ref{fig:trajs}.

\begin{table*}[t]
\setlength\extrarowheight{0.05pt}
    \centering
    \caption{Performance of Modboat controllers and foil-shaped swimmer. Modboat ranges are interquartile. \\$\dagger$ indicates stat. significance relative to Inertial, $\ddagger$ indicates stat. significance relative to thrust direction.}
    \begin{tabular}{ll|cccc|cc} \toprule
        & & \multicolumn{4}{c|}{Modboat} & \multicolumn{2}{c}{Foil-Shaped Swimmer}  \\[1pt] 
        Metric & Units & Diff. Thrust & Inertial & \textbf{Thrust Direction} & \textbf{Desat. Thrust Dir.} & Normal & Passive Tail \\ \midrule
        Body length (BL) & $\si{m}$ &  $0.15$ & $0.15$ & $0.15$ & $0.15$ &  $0.37$ ~\cite{Pollard2017AnRotor} & $0.37-0.38$ ~\cite{Pollard2019PassiveRobots} \\
        Rise time for $90^\circ$ turn & $\si{s}$ & $8.4-12\hphantom{.}$ & $4.3-6.0$ & $ 1.7\hphantom{0}-2.7^\dagger$ & $1.8\hphantom{0}-3.5^{\dagger\hphantom{\ddagger}}\hphantom{0}$ &  $\approx 27$~\cite{Pollard2019SwimmingConstraints} & \cellcolor[HTML]{000000}  \\
        Travel during $90^\circ$ turn & $\si{BL}$ & $2.4-4.1$ & $1.6-1.9$ & $0.87-1.4^\dagger$  & $0.63-1.3^{\dagger\hphantom{\ddagger}}\hphantom{0}$ &  $\approx 12$~\cite{Pollard2019SwimmingConstraints} & \cellcolor[HTML]{000000}  \\
        Rise time for $180^\circ$ turn & $\si{s}$ & Unable & $7.7-9.9$ &  $3.2\hphantom{0}-4.4^{\dagger}$ & $2.0\hphantom{0}-3.7^{\dagger\ddagger}\hphantom{0}$ & Unable~\cite{Pollard2019PassiveRobots} & $5.6-7.5$~\cite{Pollard2019PassiveRobots} \\
        Travel during $180^\circ$ turn  & $\si{BL}$  & Unable & $1.6-2.3$ & $0.50-1.1^\dagger$ & $0.38-0.69^{\dagger\ddagger}$ & Unable~\cite{Pollard2019PassiveRobots} & $0.55-1.37$~\cite{Pollard2019PassiveRobots}  \\
        RMS perpendicular error & $\si{m}$ & $0.19-0.21$ & $0.071- 0.084$ & $0.050-0.074$ & $0.027-0.073^{\dagger\hphantom{\ddagger}}$ & \multicolumn{2}{c}{\cellcolor[HTML]{000000}}\\ 
        Min. target radius & $\si{m}$ & $0.2$ & $0.2$ &  $0.050$* & $0.050$*& \multicolumn{2}{c}{\cellcolor[HTML]{000000}} \\
        Min. target distance & $\si{m}$ & $1.0$ & $1.0$ &  $0.25$* & $0.25$* & \multicolumn{2}{c}{\cellcolor[HTML]{000000}} \\ \bottomrule
    \end{tabular}
    \label{tab:comparison}
\end{table*}

The increased accuracy and maneuverability of our new method are demonstrated in Fig.~\ref{fig:trajSquare}, in which thrust direction control shows increased straightness of inter-waypoint segments and sharp turns. The inertial controller turns slightly more sharply, but requires more space and time to do so, which is highlighted in the $90^\circ$ turn characteristics and target radius in Table~\ref{tab:comparison}. The faster and shorter response of the thrust direction controller can be leveraged into tighter trajectories; this is demonstrated by the smaller square in Fig.~\ref{fig:trajSquare}, which is only $0.50\si{m}/3.3\si{BL}$ to a side and could not be achieved by inertial control.

The superiority of thrust direction control is further shown in Fig.~\ref{fig:trajLine}. Inertial control executes a very sharp turn but drifts past the target in doing so, and the maneuver induces lateral deviation on the return path. The thrust direction controller, meanwhile, turns without overshooting and maintains a straighter return trajectory. This is highlighted in the $180^\circ$ turn characteristics in Table~\ref{tab:comparison}. 

This improved performance and ability to execute fast and sharp turns culminates in allowing the Modboat --- without adjustment of the control law --- to station keep with an orbit of diameter only $1.5\si{BL}$. Neither of our previous methods could achieve this kind of maneuver, much less such a small orbit. Thrust direction control with desaturation thus outperforms prior Modboat control methods on all metrics.

We can also use Table~\ref{tab:comparison} to compare the Modboat performance to the foil-shaped robots developed by Pollard et al.~\cite{Pollard2019SwimmingConstraints}, Fedonyuk~\cite{Fedonyuk2020DynamicsFreedom}, and Lee et al.\cite{Lee2019State-feedbackVehicle}, whose controllers inspired ours. For $90^\circ$ turns convergence time is significantly faster and the distance traveled is significantly lower for the Modboat under the control laws given by~\eqref{eq:contDirTrav} and~\eqref{eq:contAdj}. This can be attributed to using flippers for propulsion rather than a foil, but it is likely also a consequence of the sinusoidal convergence term, which is faster than the integral term used by Pollard et al.~\cite{Pollard2019SwimmingConstraints}.

The Modboat also outperforms foil-shaped robots in $180^\circ$ maneuvers, as shown in Table~\ref{tab:comparison}. As a consequence of the no-slip (or small-slip) conditions at the foil tip, the foil-shaped robots cannot perform a $180^\circ$ turn in-place. The Modboat, unrestricted by such dynamic conditions, can do so  within $3.2-4.4\si{s}$ with minimal drift. Pollard et al. uses passive tail segments to reverse the direction of vortex shedding, which aids turning and allows a $180^\circ$ turn~\cite{Pollard2019PassiveRobots}. The Modboat can achieve comparable sharpness and better speed without modification of design or control technique. Moreover the Modboat's drift is almost entirely along the direction of travel when executing a $180^\circ$ turn, whereas the foil-shaped robot executes a more circular turn.


\section{Conclusion}

We present a \textbf{thrust direction} control methodology for the Modboat, a single-motor swimmer that uses conservation of angular momentum to paddle two passive flippers for thrust and steering. The oscillating limit-cycle controller was originally developed for foil-shaped robots, but we show that it is more effective at driving and steering the Modboat. This limit-cycle control, combined with a direction of travel controller, forms the novel thrust direction controller. 

On top of this, we present a novel extension to alleviate the reaction wheel problem, in which inevitable actuator saturation leads to a loss of control authority. This forms the \textbf{desaturated thrust direction} controller, which improves overall trajectory accuracy at next to no cost to large maneuvers. This allows arbitrary trajectories to be prescribed and makes this a feasible approach for driving Modboats to dock together, which, as described in our prior work, requires the top body orientation to be controlled~\cite{Knizhnik2021DockingRobot}. 

We have experimentally demonstrated that the thrust direction controller significantly outperforms prior control methods for the Modboat. This allows finer-scale maneuvers, more accurate waypoint tracking, and even station-keeping, which was impossible through previous methods. This allows the Modboat to move like a far more complex system despite being driven by only one motor and to be used for precision sensor placement or inspection tasks.

Future work includes the application of this thrust direction control law to changes in the Modboat parameters. In particular, when multiple Modboats dock together the connection significantly reduces the rotation of the top body, which may drive the controller to be unstable. Multiple Modboats swimming together is a desired use case, so the controller must remain stable under such changes. We also plan to examine the use of this controller in the presence of external flows, as well as to consider how this control might be used to take advantage of existing flow by moving between streamlines.


\section*{Acknowledgment}

We thank Dr. M. Ani Hsieh for the use of her instrumented water basin in obtaining all of the testing data.

\bibliographystyle{./bibliography/IEEEtran}
\bibliography{./bibliography/IEEEabrv,./bibliography/iros2020,./bibliography/nonpaper,./bibliography/references}

\end{document}